%% file: main.tex
\documentclass[letterpaper, 10 pt, conference]{ieeeconf}  % Comment this line out if you need a4paper

\IEEEoverridecommandlockouts                              % This command is only needed if 
\usepackage[noadjust]{cite}
\usepackage{booktabs}
\usepackage{float}
\usepackage{amsmath}
\usepackage{amsfonts}
\usepackage{graphicx}
\usepackage{subcaption}
\let\labelindent\relax
\usepackage{enumitem}
\usepackage[colorinlistoftodos,prependcaption,textsize=tiny]{todonotes}
\usepackage{gensymb}
\usepackage{hyperref}
\usepackage{lipsum}
\usepackage{balance}

\usepackage{array}
\newcolumntype{P}[1]{>{\centering\arraybackslash}p{#1}}
\newcolumntype{M}[1]{>{\centering\arraybackslash}m{#1}}

\title{\LARGE \bf
AugInsert: Learning Robust Visual-Force Policies via \\ Data Augmentation for Object Assembly Tasks
}

\author{Ryan Diaz$^{1}$, Adam Imdieke$^{1}$, Vivek Veeriah$^{2}$, Karthik Desingh$^{1}$% <-this % stops a space
\thanks{This work is supported by the University of Minnesota Undergraduate Research Opportunities Program (UROP) and the MnRI Seed Grant from the Minnesota Robotics Institute.}% <-this % stops a space
\thanks{$^{1}$University of Minnesota, Twin Cities.
        {\tt\small \{diaz0329, imdie022, kdesingh\}@umn.edu}}%
\thanks{$^{2}$Google DeepMind.
        {\tt\small vveeriah@google.com}}%
}

\let\oldtwocolumn\twocolumn
\renewcommand\twocolumn[1][]{%
    \oldtwocolumn[{#1}{
           \centering
           \includegraphics[width=0.95\textwidth]{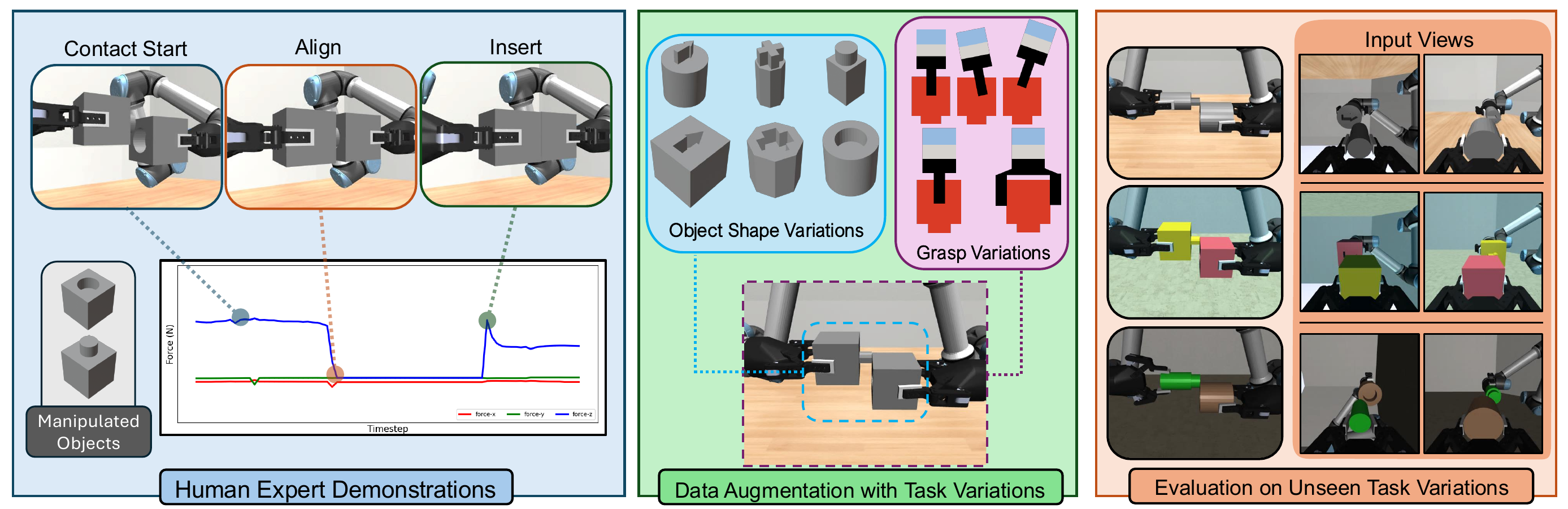}
           \captionof{figure}{\footnotesize{AugInsert is a data collection and policy evaluation pipeline aimed towards analyzing the robustness of a multisensory (vision, force-torque, and proprioception) model with respect to different observation-level task variations in object shape, grasp pose, and visual environmental appearance. Our framework introduces task variations to a dataset of human-collected demonstrations through a system of online data augmentation.}}
           \label{fig:teaser}
    }]
}

\usepackage{etoolbox}
\makeatletter
\patchcmd{\@makecaption}
  {\scshape}
  {}
  {}
  {}
\makeatother

\begin{document}

\maketitle
\thispagestyle{empty}
\pagestyle{empty}

%%%%%%%%%%%%%%%%%%%%%%%%%%%%%%%%%%%%%%%%%%%%%%%%%%%%%%%%%%%%%%%%%%%%%%%%%%%%%%%%
\begin{abstract} 

Operating in unstructured environments like households requires robotic policies that are robust to out-of-distribution conditions. Although much work has been done in evaluating robustness for visuomotor policies, the robustness evaluation of a multisensory approach that includes force-torque sensing remains largely unexplored. This work introduces a novel, factor-based evaluation framework with the goal of assessing the robustness of multisensory policies in a peg-in-hole assembly task. To this end, we develop a multisensory policy framework utilizing the Perceiver IO architecture to learn the task. We investigate which factors pose the greatest generalization challenges in object assembly and explore a simple multisensory data augmentation technique to enhance out-of-distribution performance. We provide a simulation environment enabling controlled evaluation of these factors. Our results reveal that multisensory variations such as \textit{Grasp Pose} present the most significant challenges for robustness, and naive unisensory data augmentation applied independently to each sensory modality proves insufficient to overcome them. Additionally, we find force-torque sensing to be the most informative modality for our contact-rich assembly task, with vision being the least informative. Finally, we briefly discuss supporting real-world experimental results. For additional experiments and qualitative results, we refer to the project webpage \href{https://rpm-lab-umn.github.io/auginsert/}{https://rpm-lab-umn.github.io/auginsert/}.

% \ryan{REVISIONS TO ABSTRACT: [Broader context of work: Evaluating the robustness of robotic policies to out-of-distribution environmental conditions is important for ensuring that robots can reliably operate in unstructured environments such as households. Although much work has been done in evaluating robustness for visuomotor policies, not much has been done for multisensory frameworks] [Our methods: A multisensory factor-based evaluation framework on a specific peg-in-hole task scenario; we aim to determine which factors pose the largest challenge for generalization in object assembly as well as how to increase the out-of-distribution resilience of our methods through a simple multisensory data augmentation technique. We also contribute a simulation environment that can perform controlled evaluations over these factors.][Our findings: Multisensory variations cause the most problems; unisensory data augmentation is not enough to overcome these challenges; for contact-rich assembly, force-torque is the modality that gives the most relevant information; also briefly mention real-world experiments]}

\end{abstract}

% This paper primarily focuses on learning robust visual-force policies in the context of high-precision object assembly tasks. Specifically, we focus on the \textit{contact phase} of the assembly task where both objects (peg and hole) have made contact and the objective lies in maneuvering the objects to complete the assembly. Moreover, we aim to learn contact-rich manipulation policies with multisensory inputs on limited expert data by expanding human demonstrations via online data augmentation. We develop a simulation environment with a dual-arm robot manipulator to evaluate the effect of augmented expert demonstration data. 
% Our focus is on evaluating the robustness of our model with respect to certain task variations: \textit{grasp pose, peg/hole shape, object body shape, scene appearance, camera pose,} and \textit{force-torque/proprioception noise}. We show that our proposed data augmentation method helps in learning a multisensory manipulation policy that is robust to unseen instances of these variations, particularly physical variations such as \textit{grasp pose}. Additionally, our ablative studies show the significant contribution of force-torque data to the robustness of our model. For additional experiments and qualitative results, we refer to the project webpage \href{https://bit.ly/47skWXH}{https://bit.ly/47skWXH}.

%%%%%%%%%%%%%%%%%%%%%%%%%%%%%%%%%%%%%%%%%%%%%%%%%%%%%%%%%%%%%%%%%%%%%%%%%%%%%%%%
\input{sections/01_introduction}
\label{sec:01_introduction}

\input{sections/02_related_work}
\label{sec:02_related_work}

\input{sections/03_task_setup}
\label{sec:03_task_setup}

\input{sections/04_methodology}
\label{sec:04_methodology}

\input{sections/05_experimental_setup}
\label{sec:05_experimental_setup}

\input{sections/06_experiments}

\label{sec:06_experiments}

\input{sections/07_conclusion}
\label{sec:07_conclusion}

% \section*{Acknowledgments}

% We thank Akul Mundada for his help with data collection in simulation. This project is partially funded by the Undergraduate Research Opportunities Program (UROP) at the University of Minnesota and the MnRI Seed Grant from the Minnesota Robotics Institute.

\bibliographystyle{IEEEtran} % We choose the "plain" reference style
\bibliography{references}

% \newpage

% \section*{APPENDIX}

% \input{sections/appendix}
% \label{sec:appendix}

\end{document}

%% file: sections/01_introduction.tex
\section{Introduction}
%What is the task, challenges and what are we developing as a solution?
% \ryan{TODO: Add broader context about evaluating robustness + multisensory policy learning for contact-rich tasks. Justify why peg-in-hole specifically:} 
Robust manipulation in unstructured environments requires robots to adapt to unforeseen variations in object properties, positions, and environmental conditions. This is particularly challenging for contact-rich tasks, where physical interaction plays a crucial role in successful execution. We focus on the peg-in-hole assembly task as a representative example of contact-rich manipulation. Like inserting a K-cup pod into a coffee machine, capping a bottle, or plugging in a cable, the peg-in-hole task requires precise control and adaptation to variations in contact forces and object geometries throughout the insertion process.  These shared characteristics make the peg-in-hole task a valuable benchmark for studying and improving the robustness of multisensory policies for a broader range of contact-rich manipulation tasks.
%These tasks typically involve multiple phases: a \textit{pick-up phase} where the objects (e.g., cap and bottle) are grasped and picked up by the grippers; an \textit{orienting phase}, where the objects are maneuvered into a desired relative pose before contact; and a \textit{contact phase}, where the objects are in contact, and appropriate forces are applied to complete the task (e.g., screwing the cap onto the bottle, fully inserting the plug). While the \textit{contact phase} may seem trivial to perform for humans, it poses significant challenges for robots---especially those meant to operate in household contexts---to learn from data due to several factors: a) contact-rich manipulation tasks are difficult for humans to demonstrate in order to facilitate large scale data collection for learning policies from demonstration, b) these tasks require multisensory observations (visual and tactile) and robust encoding methods to extract meaningful representations for policy learning \cite{lee2019making, chen2023visuo, spector2021insertionnet, spector2022insertionnet}, and c) humans can easily generalize these tasks to novel scenarios (e.g. varying object shapes and geometries), but such generalization is highly challenging for robot learning models while maintaining the robustness required \cite{xie2024decomposing, pumacay2024colosseum}. 

% \ryan{TODO: Rather than proposing, our goal is to evaluate ||| % }
In this paper, we address the challenge of robust manipulation in contact-rich environments by developing a multisensory policy learning framework that includes force-torque (F/T) sensing. We then experimentally evaluate how this approach can generalize to unseen task variations from limited human demonstrations. Our pipeline processes multiple camera views and F/T readings from a dual-arm setup. In an effort to increase robustness, we also explore a multisensory data augmentation method via trajectory replay that can introduce both sensor-specific (camera and F/T sensor) variations as well as physical factors such as manipulated object shape, peg and hole geometries, and grasp pose variations that affect the sensing modalities. Through this approach, we can expand small expert datasets to learn robust manipulation policies that can handle a wide variety of environmental conditions.

To analyze the robustness of our model with respect to specific observation-level task variations and understand the effect of our data augmentation method, we develop an experimental setup in the MuJoCo \cite{todorov2012mujoco} simulation environment with a dual-arm robot that can manipulate objects with peg and hole geometries to complete the assembly task. Our experiments show that certain variations, such as \textit{Grasp Pose} variations, cause large drops in success rate for our task and so should be included in training data through data augmentation in order to ensure robustness to these variations. Additionally, we conduct ablation studies to understand how each sensory modality in the multisensory setup affects the performance of the contact-rich assembly task. These studies also help identify the specific modalities impacted by each variation. We observe that touch provides the most relevant information for the task and supports model robustness; visual input, on the other hand, has the least significant impact on generalization ability while also being susceptible to many of our task variations. We provide an extensive discussion of these results in the following sections. 
The main contributions of our work are

% \begin{itemize}[leftmargin=*]
%     \item A dual-arm object assembly task formulation and a simulation environment that allows for independent application of 6 types of task variations that involve 54 different types of peg and hole objects.
%     \item An extensive set of experiments to understand the effect of data augmentation and evaluate the robustness of specific sensing modes against observation-level task variations. 
%     \item \ryan{Data augmentation is a proposed direction when trying to overcome these variations} A data augmentation pipeline integrated with the simulation environment that can generate new observations for training using a set of provided expert demonstrations through trajectory replay. 
%     % \item \textbf{Multisensory observation encoder:} A  transformer-based multisensory encoder for extracting meaningful representations to support the BC-MLP policy learner to be robust to a number of variations. %
% \end{itemize}

\begin{itemize}[leftmargin=*]
    \item A dual-arm object assembly task formulation along with a set of physical and sensor-based task variations that introduce a variety of perturbations to the environment state.
    \item A multisensory vision and force-torque policy learning framework based on the Perceiver IO \cite{jaegle2022perceiverio} architecture to perform this contact-rich task.
    \item A publicly available simulation environment for our dual-arm object assembly that supports physical and sensor-based task variations, enabling standardized benchmarking and evaluation.
    \item Extensive experiments in simulation and supplementary real-world validation analyzing the most challenging task variations for generalization in our assembly task, along with an exploration of data augmentation strategies to enhance robustness.
    % \item \ryan{TODO: Expand} An exploration of simple data augmentation techniques to enhance robustness
    % \item \kar{TODO: Expand} We conduct real-world experiments to find if the proposed data augmentation techniques improve the performance in the real-world. Basically, to observe similar trends in the problem and the proposed solution from simulation to the real-world setup.
\end{itemize}

%% file: sections/02_related_work.tex
\section{Related Work}

% Here, we will discuss notable works in three related areas of research: a) imitation learning for manipulation policy learning, b) multimodal policy learning with visual-force-torque data, c) benchmarking over-generalization capabilities of the observation encoders. \kar{I think we can swap (a) to Augmenting demonstrations area - https://genaug.github.io/}

% \subsection{Imitation Learning for Robot Manipulation Policy Learning}
% \kar{This might not be very related.}
% In recent years, there has been a surge of work in behavior cloning for robot manipulation policy learning [BC-MLP, ACT, GPT, DDPM versions]. Some also contribute to simulation environments to facilitate benchmarking [Robomimic, MimicGen, etc] and more recently [TDB] proposed a new benchmarking to evaluate how policy learning models capture diversity in the demonstration data. These methods are focused on developing BC algorithm and less on the observation representation learning. Our work assumes that our representation learning model can be plugged into the latest and greatest BC model for learning the policy. We focus on multisensory observation encoding and generalization to the task variations that are attributed by the observation space. 
% %TDB: Towards Diverse behaviors: https://arxiv.org/abs/2402.14606

\subsection{Multisensory Contact-Rich Manipulation}
Multisensory policy learning using vision and F/T data for contact-rich peg-in-hole insertion has been widely studied. Lee et al. \cite{lee2019making,lee2021detect} developed a self-supervised learning method to learn a multisensory representation using vision and F/T that can transfer across different peg and hole shapes, and Wu et al. \cite{wu2021learning} created a reward learning framework for the peg-in-hole task based on task progress. More recently, Spector et al. \cite{spector2021insertionnet, spector2022insertionnet} developed a multiview and multisensory system for localizing and performing realistic insertion tasks, Chen et al. \cite{chen2023visuo} used a transformer \cite{vaswani2017attention} encoder for vision and F/T inputs to learn a higher-quality representation, and Kohler et al. \cite{kohler2024symmetric} leveraged symmetry in the peg-in-hole task by using equivariant networks to improve sample efficiency. Although these works can achieve efficient peg-in-hole assembly, their generalization studies are limited in scope when evaluating robustness to both physical and sensory task variations. Our work aims to determine the types of variations that present the greatest challenges for generalization in this task in order to focus our efforts in increasing robustness.

% and Wu et al. \cite{wu2021learning} created a reward learning framework for the peg-in-hole task based on task progress

% , and so it is difficult to judge how well their setups would work in uncontrolled scenarios.

\subsection{Evaluating Generalization Abilities of Learned Policies}
Generalization is difficult to define in robot manipulation policy learning as there are several aspects of the robot's environment that could vary from training phase to the evaluation phase. There have been recent efforts to perform in-depth analyses of the generalization abilities of visuomotor robotic policies by decomposing task environment variations into individual variation ``factors" \cite{xing2021kitchenshift, xie2024decomposing, pumacay2024colosseum, gao2024efficient} to isolate different types of perturbations to the environment state. We aim to bring this type of analysis to the multisensory domain by introducing a set of variation factors that perturb F/T and proprioceptive inputs in addition to image inputs.

\subsection{Data Augmentation for Increased Model Robustness}
\label{sec:02c_data_augmentation}

Training on diverse demonstration datasets has allowed for the development of robot policies that can accomplish complex manipulation tasks while being robust to certain environmental variations, and a large focus has been placed on curating these datasets with minimal human intervention \cite{mandlekar2023mimicgen, jia2023seil, ankile2024juicer}. A common technique to enhance a model's robustness without requiring additional human effort in data collection is data augmentation, which involves transforming input data while preserving the original labels. This approach is most commonly used with image inputs, where transformations such as cropping, flipping, and color adjustments are applied to help the model learn invariance to these changes \cite{perez2017effectiveness, kostrikov2020image, spector2021insertionnet, zhu2022sample}. Image augmentation can also be performed at the semantic level using generative models \cite{chen2023genaug, chen2024roviaug, tang2024kalie}. However, applying this type of augmentation to contact-rich tasks poses challenges, as these tasks involve \textit{physical} variations (e.g., object size, shape), which may introduce non-independent perturbations across the multisensory input that cannot be captured through conventional offline augmentation. Our work explores an online augmentation method that can include multisensory variations in the training dataset.

% \subsection{Extrapolating Human Trajectories for Imitation Learning}

% \ryan{TODO: Probably not really relevant, replace with a section on unisensory data augmentation instead?} Imitation learning can be a powerful method for learning complex tasks in robotics, but it can be challenging to collect large enough demonstration datasets for learning effective policies. Recent works address this problem by extrapolating a small dataset of human demonstrations; \ryan{TODO: Condense this all into one argument} Mandlekar et al. \cite{mandlekar2023mimicgen} generated new trajectories by decomposing task demonstrations into object-centric subtasks, and Jia et al. \cite{jia2023seil} used different point cloud projections as new observations to simulate new transitions within a demonstration. Focusing more on robotic assembly, Ankile et al. \cite{ankile2024juicer} annotated expert trajectories with ``bottleneck" states off of which perturbations and their corresponding corrective actions could be automatically generated. These works somewhat resemble our trajectory replay method for online augmentation, but they are more suited for long-horizon tasks that do not necessarily focus on precision.

% Our task is more precise, their's is more focused on long-horizon, more complex tasks

%% file: sections/03_task_setup.tex
% Citations
%   - Ku's paper (task setup)
%   - FactorWorld and The Colosseum (task variations)

\section{Task Setup}

\begin{figure*}[h!]
     \centering     \includegraphics[width=0.95\textwidth]{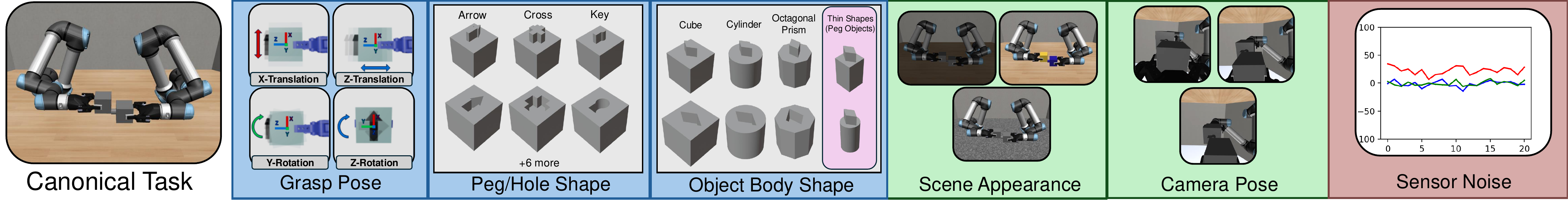}
     \caption{\footnotesize{A visualization of each of the task variations used in our environment setup. We differentiate between physical task variations (in blue) and sensor-based task variations that target vision (green) and force-torque/proprioception (red).}}
     \label{fig:task_variations_visual}
\end{figure*}

\subsection{Assembly Task Definition}
Our experimental setup consists of a dual-arm robot manipulator with a multisensory configuration, featuring two F/T sensors and two RGB cameras attached to its wrists. We elect to use a dual-arm setup to emulate object assembly scenarios in which neither the peg nor hole are fixed to the environment, unlike the single-arm peg-in-hole setups of previous work. The robot is tasked with performing an insertion assembly, where one arm's gripper holds a peg-shaped object and inserts it into a hole-shaped object held by the other arm's gripper. Since our focus is on the contact phase of the assembly, the objects are already in contact at the start of the task. 

The objective of the robot learning framework is to execute the assembly task without explicit information about the object geometries or peg and hole shapes, while maintaining robustness to various task variations. We take the behavior cloning approach where expert demonstrations are used to train the contact-rich manipulation policy to perform the assembly.

\subsection{Task Initialization}

% \ryan{TODO: Revise} Our task employs a dual-arm setup, with one arm holding the peg and another holding the object with a hole. Both are parallel to the ground. We assume that the peg starts in contact with the other object and that the peg and hole shapes are rotationally aligned. 

% The task is initialized such that the peg and the hole are offset within the range of \ryan{[LOWER, UPPER]} for both the X and Z axes. 
% Our task is setup such that while the arm holding the peg is moving, the other arm stays compliant and applies a constant amount of force until the peg and hole are aligned. The task is considered successful if the difference in position of the two objects relative to each other falls below a certain threshold corresponding to a successful insertion.

The task is initialized with the peg and hole offset within a range of [1.5cm, 3.0cm] along both the X and Y axes relative to the object coordinate frame (perpendicular to the direction of insertion).
% The goal of the task is to align the peg and hole, and then insert the peg into the hole. 
Our setup ensures that while the arm holding the peg moves, the other arm remains compliant, constantly applying a force until the peg and hole are aligned. To define a successful task rollout, we consider position coordinates $\mathbf{p} = (p_x, p_y, p_z)$ for the peg object and $\mathbf{h} = (h_x, h_y, h_z)$ for the hole object in the global coordinate frame. We set thresholds $\mathbf{d} = (d_x, d_y, d_z)$ such that during a successful insertion, $|p_i - h_i| < d_i$ for all axes $i \in \{x,y,z\}$.

\subsection{Task Variations}

To evaluate the robustness of our trained models, we design a set of observation-level task variations
%inspired by \cite{xie2024decomposing} and \cite{pumacay2024colosseum}
which alter the distribution of incoming observations while preserving the underlying task. In total, there are six variations that are part of the experiments (see Figure \ref{fig:task_variations_visual} for sample visualizations of these variations):
% To evaluate the robustness of our trained models, we design a set of observation-level task variations inspired by \cite{xie2024decomposing} \cite{pumacay2024colosseum} that change the distribution of incoming observations while maintaining the underlying task. In total there are 6 variations (See Figure.~\ref{fig:task_variations_visual} for visualization of these variation s):

\begin{enumerate} [leftmargin=*]
    \item \textbf{\textit{Peg and Hole Shape:}} There are 9 possible peg and hole shapes: arrow, circle, cross, diamond, hexagon, key, line, pentagon, and u. The peg and hole in a given pair share the same shape and are built with a 5mm tolerance to ensure insertion compatibility.
    \item \textbf{\textit{Object Body Shape:}} There are 3 possible object body shapes (cube, cylinder, and octagonal prism), and the peg and hole in a given pair may or may not share the same body shape. Additionally, we create thinner versions of the peg of 60\% width to introduce variability, resulting in 6 total peg and hole object pairs.
    \item \textbf{\textit{Grasp Pose:}} Our grasp pose variation follows the approach in \cite{ku2024evaluating}. X-axis translation offsets lie within [-1.7cm, 1.7cm] of the object's center (scaled down to $60\%$ for thin object body shapes). Z-axis translation offsets lie within [0.0cm, 1.4cm] of the object's center. Y-axis rotation offsets range from [-10.0$\degree$, 10.0$\degree$], and Z-axis rotation offsets are sampled from \{0$\degree$, 90$\degree$, 180$\degree$, 270$\degree$\}.
    % Translations vary in \ryan{[LOWER, UPPER]} and rotations in \ryan{[LOWER, UPPER]}, with these ranges scaled down for smaller object body shapes to maintain physically plausible grasps.
    \item \textbf{\textit{Scene Appearance:}} This category encompasses variations in lighting, floor texture, and object color. 
    \item \textbf{\textit{Camera Pose:}} The wrist camera positions are perturbed in the range of [-4cm, 4cm] on each axis, and camera orientations are perturbed by a random axis-angle rotation. The axis is sampled from a standard Gaussian and the angle is sampled from [0$\degree$, 5$\degree$].
    \item \textbf{\textit{Sensor Noise:}} We add zero-mean Gaussian noise to low-dimensional measurements, including both force-torque and proprioception readings for both arms. The standard deviations of sampled noise for each measurement are 5N for force, 0.15N-m for torque, 0.1cm for end-effector position on each axis, and 0.57$\degree$ for end-effector orientation on each axis.
    %with standard deviations set to approximately $5\%$ of the maximum measurement for force-torque and $4\%$ of the maximum task-initialized offset for proprioception.
\end{enumerate}

% \begin{enumerate} 
%     \item \textbf{Peg and Hole Shape:} There are 9 possible peg and hole shapes, with the peg and the hole sharing the same shape to ensure insertion compatibility.
%     \item \textbf{Object Body Shape:} There are 3 possible object body shapes, and the two objects may not necessarily share the same shape. We also create thinner versions of these shapes for the peg object to introduce additional variability.
%     \item \textbf{Grasp Variations:} Our grasp variation formulation is similar to \cite{ku2024evaluating}, and we refer to that paper for a more detailed overview. Translations vary in \ryan{[LOWER, UPPER]} and rotations in \ryan{[LOWER, UPPER]}, with these ranges scaled down for smaller object body shapes to maintain physically plausible grasps.
%     \item \textbf{Scene Appearance:} This category encapsulates a composition of lighting, floor texture, and object color variations.
%     \item \textbf{Camera Angle:} We vary the position (in the range of \ryan{[LOWER, UPPER]} for all axes) and orientation (in the range of \ryan{[LOWER, UPPER]} for all axes) of the two wrist cameras.
%     \item \textbf{Sensor Noise:} We add zero-mean Gaussian noise to low-dimensional measurements with standard deviations equal to approximately $5\%$ of the maximum measurement for force-torque, and \ryan{$5\%$?} of the maximum task-initialized offset for proprioception.
% \end{enumerate}

\noindent \textbf{Canonical Task Setup:} We define a ``canonical" task setup which represents an environment without any task variations applied. For discrete task variations, we choose the \texttt{key} peg and hole shape, \texttt{cube} object body shape, and \texttt{light-wood} floor texture in our canonical setup in simulation. We also construct a similar setup in the real world, discussed further in Section \ref{sec:04d_real_world_setup}.

%% file: sections/04_methodology.tex
% Citations
% [Imitation Learning Framework]
%   - Robomimic (for BC-MLP)
%   - VTT (for encoder)
%   - PerceiverIO (for encoder)
%   - PerAct (for PerceiverIO inspiration)
% [Data Collection (Sim Environment)]
%   - Robosuite
%   - MuJoCo
%   - MimicGen, SEIL (for trajectory cloning)
\section{Methodology}

\begin{figure*}[h!]
     \centering
     \includegraphics[width=0.9\textwidth]{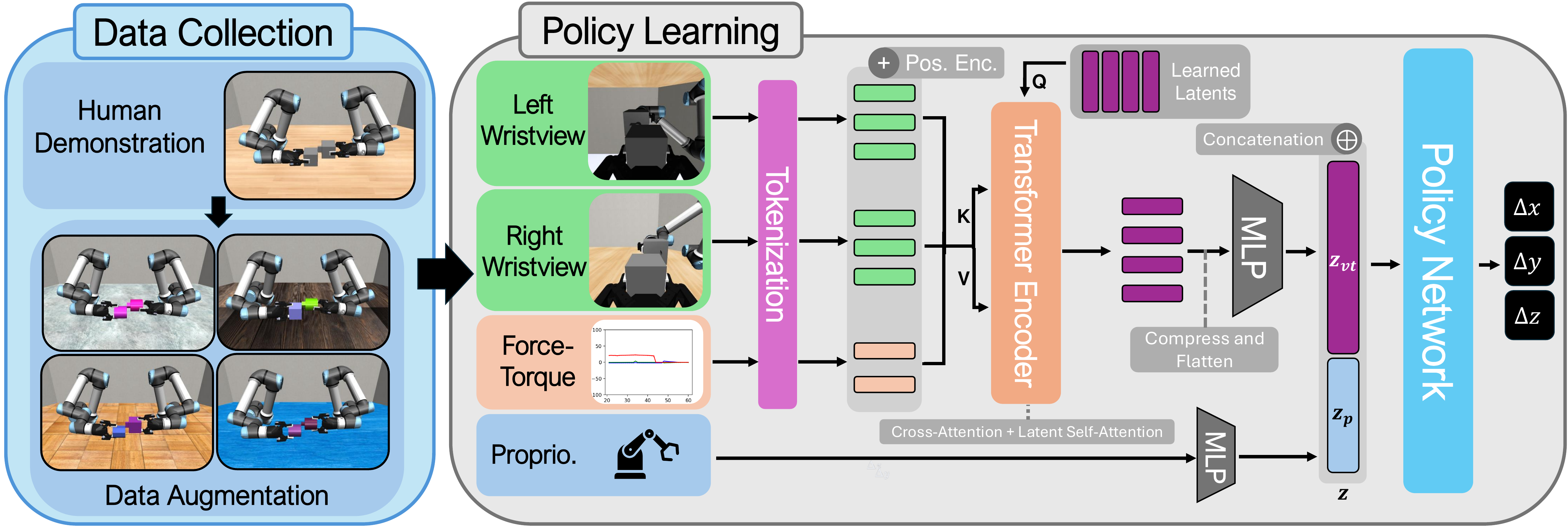}
     \caption{\footnotesize{An overview of our data collection and policy learning framework. We use BC-MLP \cite{mandlekar2022matters} with a multilayer perceptron policy network to output actions. Image and force-torque observations are encoded with a visuotactile transformer \cite{chen2023visuo} that includes a cross-attention step with a set of learned latent vectors (similar to Perceiver IO \cite{jaegle2021perceiver, jaegle2022perceiverio}). More details on our network architecture can be found in our supplementary material and website.}} 
     \label{fig:architecture}
\end{figure*}

\subsection{Imitation Learning Framework}
\textbf{Observation and Action Spaces:} In our task, the observation space is defined as a composition of four modality spaces, $\mathcal{O} = \mathcal{I}_{left} \times \mathcal{I}_{right} \times \mathcal{T} \times \mathcal{S}$. The image spaces, $\mathcal{I}_{left} \subseteq \mathbb{R}^{84\times84\times3}$ and $\mathcal{I}_{right} \subseteq \mathbb{R}^{84\times84\times3}$, represent $84\times84$ RGB wrist views from the left and right arms. The tactile space, $\mathcal{T} \subseteq \mathbb{R}^{32\times12}$, corresponds to a history of the last 32 force and torque readings from both arms (concatenated), while the robot proprioceptive state space, $\mathcal{S} \subseteq \mathbb{R}^{14}$, represents the absolute end-effector positions and orientations (expressed as quaternions) for both arms (concatenated). Our action space, $\mathcal{A} = [0,1]^3$, consists of end-effector position deltas relative to the current pose.

\textbf{Policy Learning:} The goal is to learn a policy $\pi: \mathcal{O} \rightarrow \mathcal{A}$ that maps observations to actions, enabling task completion. In the imitation learning setting, an expert policy $\pi^*$ is provided, where $a^* = \pi^*(o)$ represents the optimal action for an observation $o \in \mathcal{O}$. Our objective is to learn a policy $\pi$ that closely resembles the behavior of $\pi^*$. There are several approaches to learning such a policy from demonstrations, with the simplest being \textit{behavior cloning}. 
In behavior cloning, the expert provides a dataset of $N$ demonstration trajectories $\mathcal{D} = \{\{(o_i, a^*_i)\}_{i=1}^{n_j}\}_{j=1}^N$, where $n_j$ is the horizon for demonstration $j$. 
The policy $\pi$ is then trained to replicate the expert actions from $\pi^*$ for the corresponding observations using supervised learning. Our observation encoder and policy network (shown in Figure \ref{fig:architecture}) are trained end-to-end using an $L_2$ loss between expert and predicted actions. 

\subsection{Data Collection with Human Experts}
% Detail simulation environment
% Human demonstrations on canonical rollouts, replaying trajectories on rollouts with different variations applied
% Trajectory cloning: Online observation augmentations with multimodal consistency (with variations such as grasp variations, peg/hole shape variations, body shape variations)

We collect a dataset of 50 human demonstrations in our simulation environment, built using the Robosuite framework \cite{zhu2020robosuite} with MuJoCo \cite{todorov2012mujoco} as the simulation engine. All demonstrations are performed in the canonical environment setup described in Section \ref{sec:03_task_setup}. A human expert teleoperates the robot's moving arm via keyboard inputs, with actions recorded as the difference between the end-effector positions in consecutive frames. We also collect 50 human expert demonstrations in our real-world environment (see Section \ref{sec:04d_real_world_setup}), using a joystick teleoperation setup for data collection.
% The simulation automatically terminates and records the demonstration upon detecting a successful completion.
%\kar{In the real-world experiments, expert demonstrations are similarly collected through teleoperation.}

\subsection{Multisensory Data Augmentation}

% A common technique to enhance a model's robustness without requiring additional human effort in data collection is data augmentation, which involves transforming input data while preserving the original labels. This approach is most commonly used with image inputs, where transformations such as cropping, flipping, and color adjustments are applied to help the model learn invariance to these changes \cite{perez2017effectiveness, spector2021insertionnet, zhu2022sample}. Image augmentation can also be performed at the semantic level using generative models \cite{chen2023genaug}. However, applying this type of augmentation to contact-rich tasks poses challenges, as these tasks involve \textit{physical} variations (e.g., object size, shape), which may introduce non-independent perturbations across the multisensory input that cannot be captured through conventional offline augmentation. 

In Section \ref{sec:02c_data_augmentation}, we hypothesized that \textit{offline} data augmentation methods may not be effective for increasing robustness to multisensory variations. To address this, we explore an \textit{online} data augmentation technique by replaying human-generated trajectories on task instances with identical initial object positions and orientations but with a subset of task variations applied. Given the previously-defined dataset of expert demonstration trajectories $\mathcal{D}$, task variations $\mathcal{V}$ (e.g. \textit{Grasp Pose}, \textit{Peg/Hole Shape}, etc.), and a function $f_{\mathcal{K}}: \mathcal{O} \rightarrow \mathcal{O}$ that returns an input observation with a subset $\mathcal{K} \subseteq \mathcal{V}$ of task variations applied, our online augmentation process takes an expert demonstration $d_j = \{(o_i, a_i^*)\}_{i=1}^{n_j} \in \mathcal{D}$ and outputs a set of new demonstrations $\Omega_{d_j} = \{\{(f_{\mathcal{K}}^{t}(o_i), a_i^*)\}_{i=1}^{n_j}\}_{t=1}^T$, where $T$ is the number of augmentations per expert demonstration. The indexed functions $f_{\mathcal{K}}^1, \dots, f_{\mathcal{K}}^T$ indicate that although each application of task variations $\mathcal{K}$ is different between each generated demonstration in $\Omega_{d_j}$, the specific application of $\mathcal{K}$ is consistent for each observation in a given augmented demonstration. After the augmentation process, we construct a new dataset $\hat{\mathcal{D}} = \mathcal{D} \cup \left( \bigcup_{n=1}^N \Omega_{d_n} \right)$ that can be used for training. It can be noted that in simulation, this data augmentation approach resembles domain randomization \cite{tobin2017domain, mehta2020active} by randomizing the simulator state, but our focus here is on variations for a multisensory setup as opposed to a camera-only one.

% Recent works [MimicGen, SEIL, others?] attempt to minimize human effort during data collection by extrapolating human-generated trajectories to create new expert trajectories that can be appended onto the dataset, but they operate on more complex tasks with longer horizons. Our trajectory cloning framework does not produce entirely new trajectories, but we have found that the relatively low number of human demonstrations collected is sufficient for learning the base task (detailed in the following sections).

\subsection{Real World Setup}
\label{sec:04d_real_world_setup}

\begin{figure}[h!]
     \centering     \includegraphics[width=1.0\columnwidth]{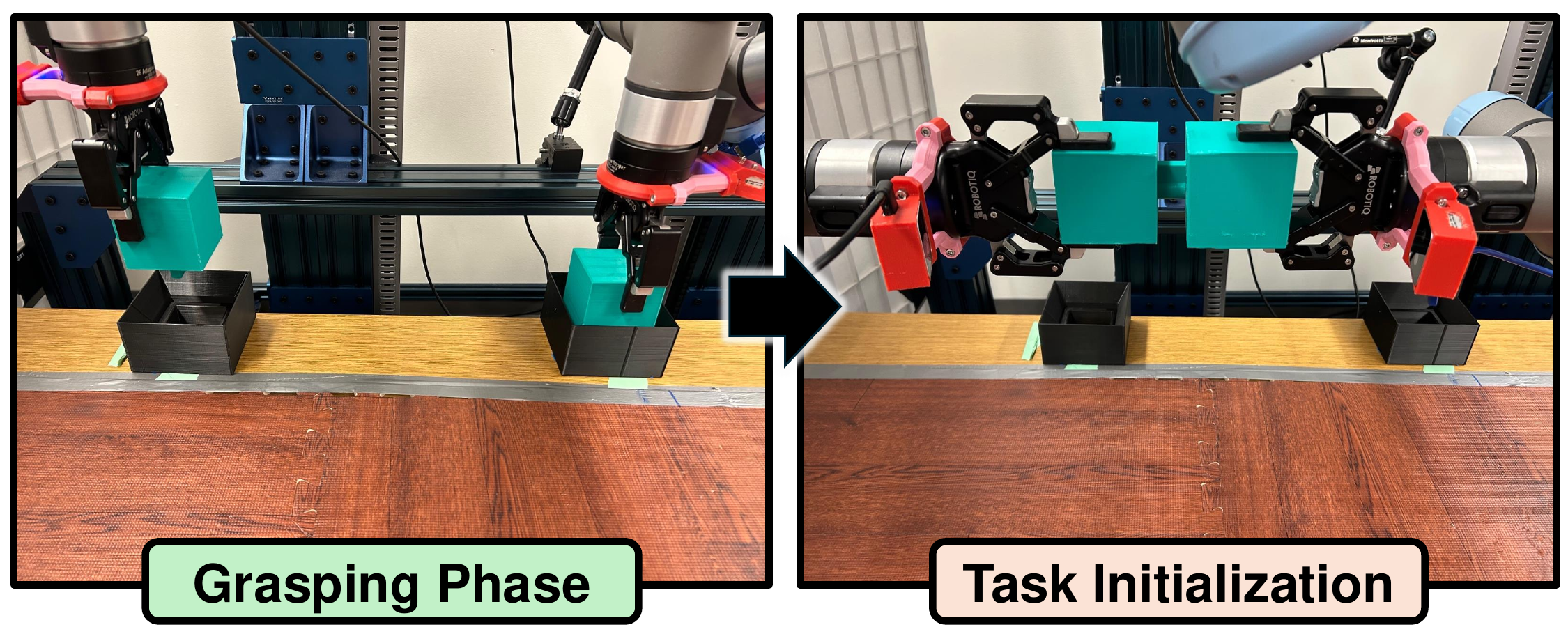}
     \caption{\footnotesize{Our real-world experimental setup. Peg and hole object models are 3D-printed directly from the model files used in our simulation environment. We implement an automated grasping phase that grasps the peg and hole objects from a set of fixtures while optionally introducing random perturbations to grasp poses. This phase is then followed by the task initialization phase for real-world experiments.}}
     % We implement a grasping phase between task executions in the real world pipeline to allow for swapping out objects and precise grasp pose changes.
     \label{fig:real-world-setup}
\end{figure}

Our real-world task setup, shown in Figure \ref{fig:real-world-setup}, is built to mirror our simulation setup as closely as possible. The setup consists of two UR5e arms with Robotiq 2F-85 grippers; a Realsense D405 RGB camera is mounted on the wrist of each arm. We designate one arm to be compliant, applying a constant amount of force while the other arm moves according to the actions given to it by the policy. In contrast to policies trained in simulation, our real-world policies predict 2-dimensional delta actions in the axes perpendicular to the axis of insertion (rather than 3-dimensional actions that include the axis along the direction of insertion), in order to prevent potentially unsafe interactions that may occur as a result of a premature insertion attempt. Once the peg and hole are aligned, the compliant arm automatically moves its held object forward to complete the insertion. To facilitate precise grasp pose generation and allow for efficient object swapping between rollouts, we insert an intermediate grasping phase between each task execution in which both arms can place and pick up the peg and hole objects from a set of fixtures.

%% file: sections/05_experimental_setup.tex
\section{Experimental Setup}

We conduct experiments to evaluate the robustness of our model with respect to each of the 6 implemented task variations (see Section \ref{sec:03_task_setup}), as well as a combination of all variations (denoted as \textit{All Variations}). 

\subsection{Training and Evaluation Details}

In all experiments, we train the models for 25000 training steps total. In simulation, we draw inspiration from Robomimic \cite{mandlekar2022matters} and perform 50 rollouts every 1250 training steps on the same task variations used in the training dataset. The model checkpoint that achieves the highest success rate in these rollouts during training is selected for evaluation on unseen scenarios. This training process is conducted over 6 random seeds per model, and the performance is averaged across all seeds during evaluation. A rollout is considered successful if it results in a successful insertion, and it is deemed failed if the maximum horizon is exceeded without insertion. Additionally, a rollout fails if the force-torque measurement surpasses a predefined threshold (100N of force or 6N-m of torque in simulation), to prevent unsafe behavior that could damage the robot arms or objects.

We train our real-world models with the same hyperparameters as those in simulation, although we only initiate 1 training seed per model (rather than 6). Additionally, we evaluate each model at the end of the entire training process, rather than performing rollouts during the training process to determine the best checkpoint. Success rates are determined by the number of successful trials over 20 rollouts. Successes and failures follow the same general criteria as in simulation, though a human manually annotates successes and failures per trial.

% In all experiments, we train models for 100 epochs 
% %(250 gradient steps per epoch) 
% and perform 50 rollouts on the same task variations that were collected in the training dataset. The model checkpoint that achieves the highest success rate in these rollouts throughout the training process is then selected for evaluation. This training process is conducted over 6 seeds for each model, and we average the performance over all seeds during evaluation. During evaluation, a rollout is considered successful if it achieves a successful insertion, and is considered failed if it passes the maximum horizon without achieving insertion. A rollout is also failed if the force-torque measurement surpass a certain threshold during evaluation; this is done in order to avoid rewarding dangerous behavior that could potentially damage the arms and/or objects.

\subsection{Evaluation on Unseen Task Variation Instances}
\label{sec:05A_experimental-setup}

To ensure rigorous evaluation, we explicitly separate task variation instances encountered during training from those used for evaluation within each variation category. For instance, when training with \textit{Grasp Pose} variations in simulation, we include demonstrations involving x-axis translation and z-axis translation and rotation, while reserving y-axis rotation for evaluation. This approach guarantees that the model encounters unseen variations during evaluation, enabling us to assess its generalization to out-of-distribution inputs across all variation categories. A detailed overview of training versus evaluation instances for each variation category is provided in Table~\ref{table:train_eval_instances}.

% \begin{table}[h]
% \centering
% \begin{tabular}{l|p{2.3cm}|p{2.3cm}}

% \toprule
%  \multicolumn{1}{c|}{\textbf{Task Variation}} & \multicolumn{1}{c|}{\textbf{Train Instances}} & \multicolumn{1}{c}{\textbf{Eval Instances}} \\ \midrule
%     \textit{Grasp Pose} \cite{ku2024evaluating} & XT, ZT, ZR & XT, ZT, ZR, YR \\ & {\color{blue}XT, ZT} & {\color{blue}XT, ZT, ZR} \\ \midrule
%     \textit{Peg/Hole Shape} & key, circle, cross & arrow, u, pentagon, line, hexagon, diamond \\ & {\color{blue}key} & {\color{blue}cross} \\ \midrule
%     \textit{Object Body Shape} & \textbf{Peg/Hole Objects:} cube, cylinder & \textbf{Hole Object}: cube, cylinder, octagonal prism, \textbf{Peg Object}: thin cube, cylinder, and octagonal prism \\ &{\color{blue}cube} & {\color{blue}cube, cylinder} \\ \midrule
%     \textit{Scene Appearance} & 6 floor textures, object color & \thinspace14 unseen floor textures, object color, lighting \\ & {\color{blue}1 floor texture} & {\color{blue}1 unseen floor texture, object color, lighting}\\
%  \bottomrule
% \end{tabular}
%  \caption{\footnotesize{Instances for task variations during training (if included the training set) and evaluation in simulation (shown in black) and real world (shown in {\color{blue}blue}) environments. Task variations not in this table are the same both in training and evaluation.}}
%  \label{table:train_eval_instances}
% \end{table}

\begin{table}[h]
\centering
\begin{tabular}{l|p{2.3cm}|p{2.3cm}}

\toprule
 \multicolumn{1}{c|}{\textbf{Task Variation}} & \multicolumn{1}{c|}{\textbf{Train Instances}} & \multicolumn{1}{c}{\textbf{Eval Instances}} \\ \midrule
    \textit{Grasp Pose} \cite{ku2024evaluating} & XT, ZT, ZR & XT, ZT, ZR, YR \\ \midrule
    \textit{Peg/Hole Shape} & key, circle, cross & arrow, u, pentagon, line, hexagon, diamond \\ \midrule
    \textit{Object Body Shape} & \textbf{Peg/Hole Objects:} cube, cylinder & \textbf{Hole Object}: cube, cylinder, octagonal prism, \textbf{Peg Object}: thin cube, cylinder, and octagonal prism \\ \midrule
    \textit{Scene Appearance} & 6 floor textures, object color & \thinspace14 unseen floor textures, object color, lighting \\
 \bottomrule
\end{tabular}
 \caption{\footnotesize{Instances for task variations during training (if included in the training set) and evaluation. Task variations not in this table are the same both in training and evaluation.}}
 \label{table:train_eval_instances}
\end{table}

%% file: sections/06_experiments.tex
\section{Experiments and Results}
In this section, we address a series of questions related to generalization through experimental evaluations.

\subsection{Which task variations are most difficult to generalize to?}

\label{sec:05A_task-variation-difficulty}

\begin{figure}[h!]
     \centering
     \includegraphics[width=1.0\columnwidth]{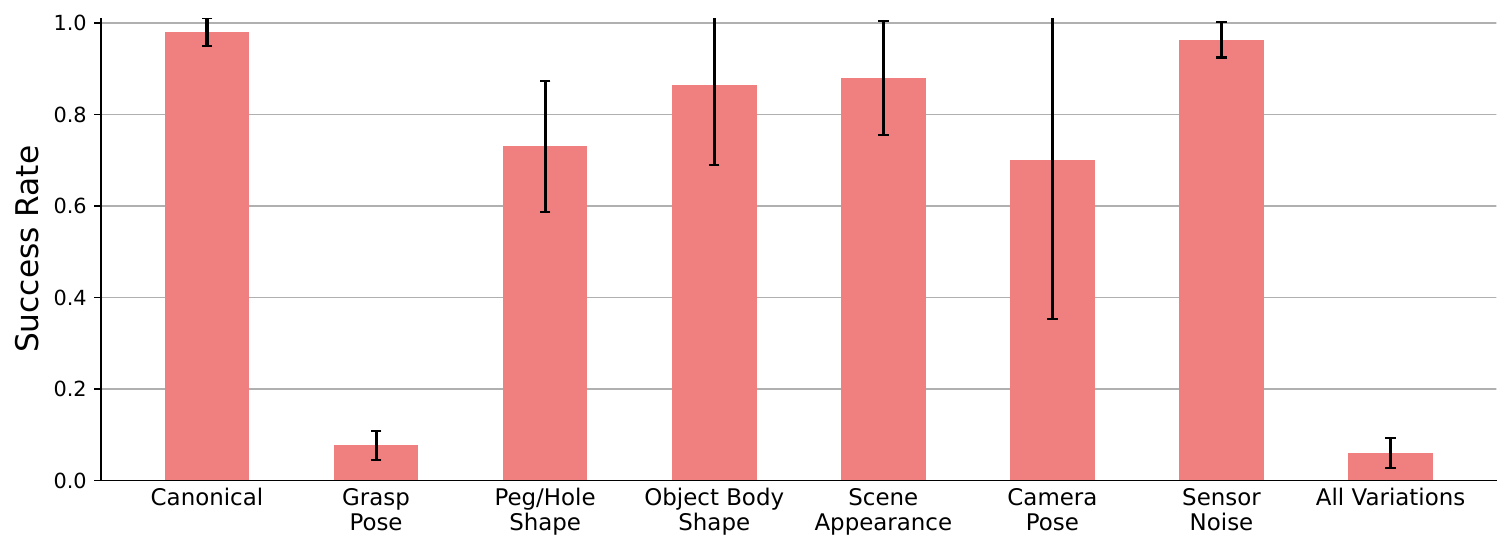}
     \caption{\footnotesize{Success rates on each task variation for a model trained exclusively on non-augmented human demonstration data. Error bars represent one standard deviation from the mean. The model suffers the largest success rate drop compared to the canonical environment when evaluated on \textit{Grasp Pose} variations, and subsequently does poorly when evaluated on \textit{All Variations}.}}
     \label{fig:task_variations_eval}
\end{figure}

In determining the difficulty of generalizing to each of our task variations, we train a model exclusively on human-collected demonstrations without any task variations applied and report the success rate for each task variation during evaluation. These results can be found in Figure \ref{fig:task_variations_eval}.

% Grasp variations perturbs all modalities in a significant manner
% Peg/hole shape drop might come from both the visual perturbation but also the different precision of insertion required for each shape (circle vs. arrow), which the human demonstrator cannot take into account when demonstrating on just the canonical setup

\noindent \textbf{Takeaways:} We observe that \textit{Grasp Pose} variations pose the greatest challenge for generalization out of all of the individual task variations, as we see a drop from a mean success rate of \textbf{0.987} on \textit{Canonical} rollouts with no task variations to \textbf{0.087} when \textit{Grasp Pose} variations are applied. We hypothesize that the large negative impact on performance comes from the significant perturbation that grasp variations apply to all sensing modalities, unlike other variations such as \textit{Scene Appearance} and \textit{Sensor Noise} which only target specific modalities and thus have a smaller negative impact for the overall model. 

\subsection{Which task variations included in the training set produce the largest impact on robustness?}

\begin{figure}[h!]
     \centering
     \includegraphics[width=1.0\columnwidth]{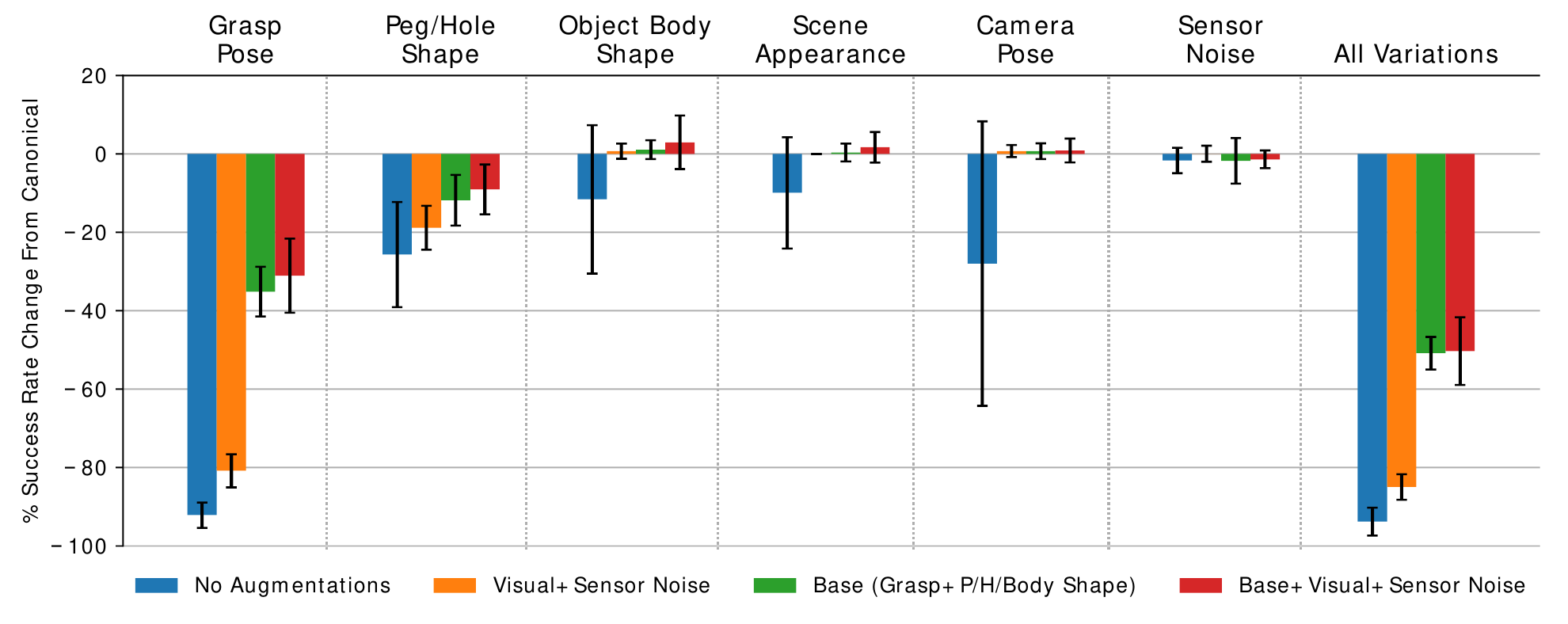}
     \caption{\footnotesize{\% success rate changes on each task variation for models trained on different subsets of task variations. Error bars represent one standard deviation from the mean. The addition of grasp variations to the training set greatly improves generalization to unseen grasp variations, while visual variations and sensor noise do not have any significant effect on generalization ability on physical task variations.}}
     \label{fig:training-set-evals}
\end{figure}

To evaluate the effect of introducing task variations to the training dataset, we evaluate models trained on datasets augmented with different subsets of our task variations. These datasets contain the original collected human demonstrations as well as 6 augmentations per demonstration, with each augmentation containing a composition of a subset of task variations (which we refer to as the ``training set variations" for that specific dataset). We evaluate these models both on instances of their training set variations that were unseen during training (as discussed in Section \ref{sec:05A_experimental-setup}) as well as all variations not included in the training set (which we refer to as the ``evaluation set variations" for that specific dataset). In Figure \ref{fig:training-set-evals} we report \% success rate change from the canonical environment success rate averaged over 6 seeds for each variation. Here, we define \% success rate change as 

\[\text{\% success rate change} = \frac{\text{task var. success} - \text{canon. success}}{\text{canon. success}}\]

% Pretty self-explanatory, when you add data to the training set, the model will learn to deal with that type of data. Adding more variations does not improve performance on those variations as those variations already had high success rates with the canonical policy. We see slight decreases in success rates in variations such as sensor noise, scene appearance, and the canonical task when adding variations to the training set; the model has to learn to deal with more complex patterns in the dataset, so it is less equipped to deal with the simpler variations. 

% See an interesting drop in generalization gap for visual vars when grasp vars are added to the dataset (but not visual vars); this can be attributed to the effect of grasp variations on the wrist-views being similar to directly changing the camera angle, as the view of both the held and opposite objects will be significantly changed (especially with rotations)

\noindent \textbf{Takeaways:} We observe that including only unisensory variations such as \textit{Scene Appearance} (visual variations) and \textit{Sensor Noise} do not significantly improve policy robustness with respect to multisensory variations such as \textit{Grasp Pose}, signaling that unisensory data augmentation cannot handle multisensory perturbations. Although there is a limited amount of visual variations in this dataset, supplemental experiments showed that performance was not improved even with a greatly expanded set of visual and sensor noise variations. On the other hand, we observe that the additions of \textit{Peg/Hole Shape}, \textit{Object Body Shape}, and \textit{Grasp Pose} (considered the \texttt{Base} variations for this task) to the training set greatly reduce the generalization gap on unseen instances of these variations during evaluation. Curiously, we also observe a reduced generalization gap for a dataset with the \texttt{Base} training set variations on the evaluation set variations of \textit{Scene Appearance} and \textit{Camera Pose}, even though these variations had not been explicitly included in the dataset. This may be due to the similarity between the effects of applying grasp variations and perturbing the camera pose, as both alter the view of the object held by the gripper and the opposing object. Additionally, the resulting visual variations may have contributed to improving the model's robustness to changes in scene appearance. Explicitly adding both visual variations (\textit{Scene Appearance} and \textit{Camera Pose}) and \textit{Sensor Noise} to the training set does not further enhance generalization to their respective variations during evaluation. 

\subsection{Can increasing the number of augmentations per demonstration improve robustness?}

\begin{figure}[h!]
     \centering
     \includegraphics[width=1.0\columnwidth]{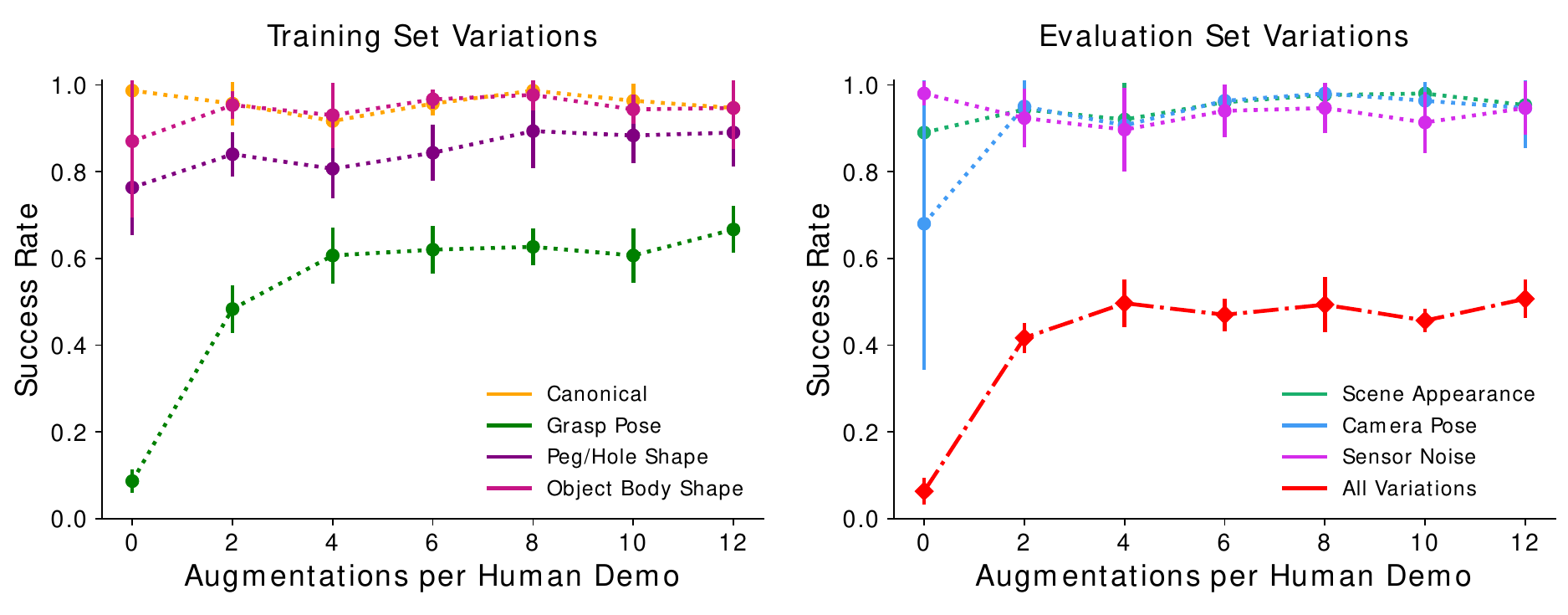}
     \caption{\footnotesize{Success rates on each task variation for models trained on a base set of training variations with different number of augmentations for each human demonstration. \textit{All Variations} represents a composition of both training set and evaluation set variations. Error bars represent one standard deviation from the mean. Success rate on \textit{Grasp Pose} variations increases the most with an increasing number of augmentations, while the other task variations maintain stable success rates.}}
     \label{fig:num-clones}
\end{figure}

Building off of our investigation into determining the ideal training set variations, we also seek to analyze the effect of adding more augmentations per demonstration involving these variations. Aligning with the previous experiment, we choose \textit{Grasp Pose}, \textit{Peg/Hole Shape}, and \textit{Object Body Shape} as our training set variations, and train models on datasets with different numbers of augmentations per human demonstration. Success rates on each task variation are reported in Figure \ref{fig:num-clones}.

% Most significant increase in grasp variations (and all variations because of grasp variations), slight increase in peg and hole shape and object body shape per trajectory clones but improvement is not significant. Peg/hole/body shape are discrete variations and so adding more augmentations may not be as beneficial for these variations due to redundant data; grasp variations is a continuous variation which benefits more from a larger sample

\noindent \textbf{Takeaways:} The most significant improvement in success rate as the number of augmentations increases seems to be in the performance on \textit{Grasp Pose} evaluations, with a more subtle upward trend in the other training set variations. Since the task variations of \textit{Peg/Hole Shape} and \textit{Object Body Shape} are discrete variations with a small subset of all possible shapes being included in the training set, they would benefit less from having more augmentations, as the dataset would start to contain redundant instances of these variations. \textit{Grasp Pose} variations, on the other hand, are continuous and so would benefit more from a larger sample of grasps. For the evaluation set variations (aside from \textit{All Variations}), the success rate remains stable, suggesting that the model is not overfitting to the training set variations even when the dataset has more samples biased towards those perturbations.

\subsection{How much does each sensory modality contribute to model robustness?}
\label{sec:05E_input-modality-combinations}

\begin{figure}[h]
     \centering
     \includegraphics[width=0.95\columnwidth]{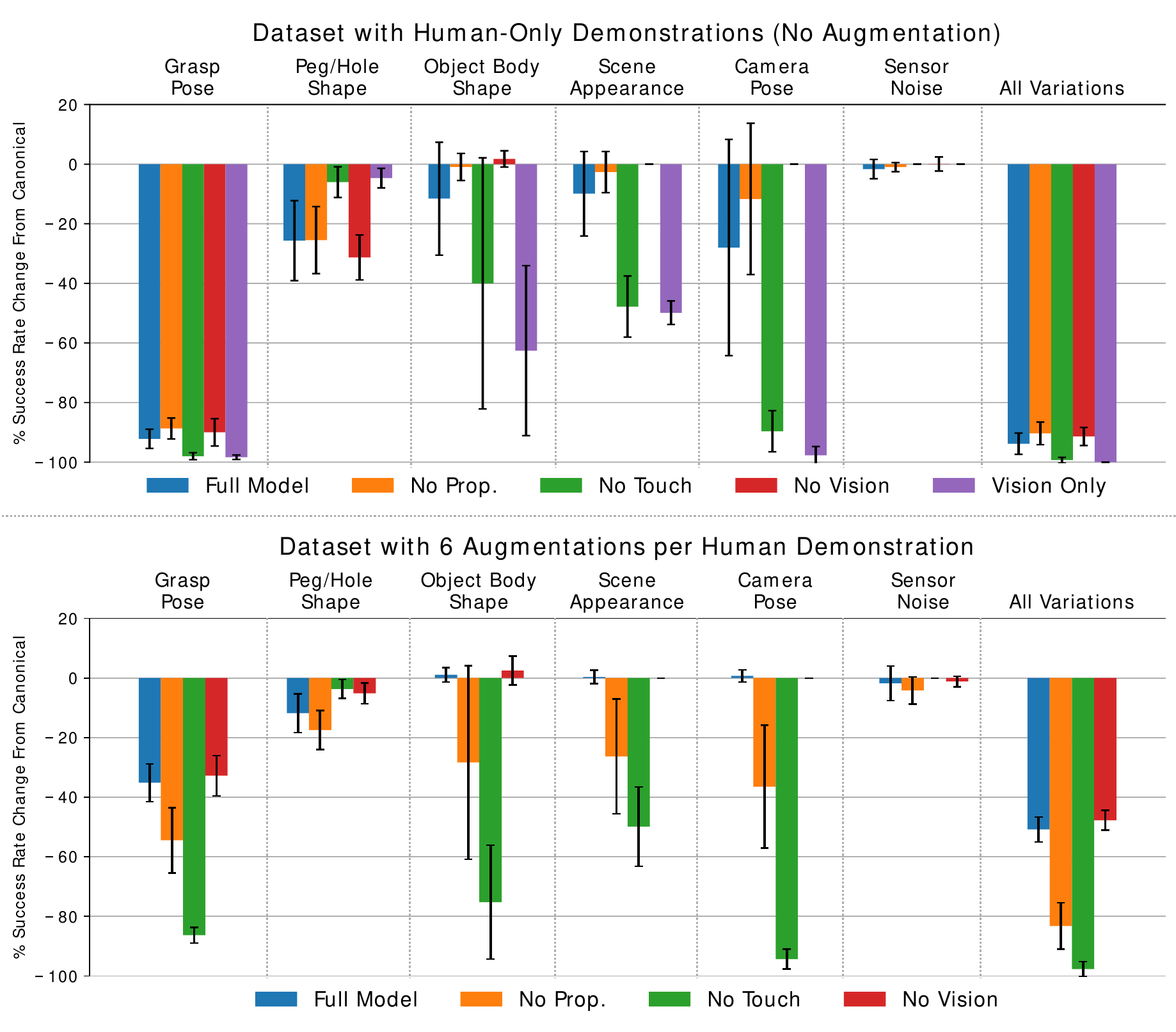}
     \caption{\footnotesize{
     \% success rate changes on each task variation for models with different modality input combinations trained on no task variations (top) or a subset of task variations (bottom). The \texttt{Vision Only} model is omitted from the bottom plot due to training instability. Error bars represent one standard deviation from the mean. The removal of force-torque input sees the largest \% success rate drop for many of the task variations out of each of the individual modalities, while the removal of vision has little impact on \% success rate change compared to the full model.}}
     \label{fig:modality-input-ablation}
\end{figure}

In an effort to investigate the significance of each of the modalities in our system---vision (wristview cameras), touch (force-torque), and proprioception---we conduct an ablation study with models that have one or more input modalities missing. We evaluate each model when trained on a dataset with no variations (i.e. no augmentations) and a dataset with 6 augmentations per demonstration on a training variation set of \textit{Grasp Pose}, \textit{Peg/Hole Shape}, and \textit{Object Body Shape} to analyze how each modality combination reacts when task variations are introduced during training. Figure \ref{fig:modality-input-ablation} shows reported \% success rate change from the canonical environment success rate averaged over 6 seeds for each variation.

\noindent \textbf{Takeaways:} We observe that out of the individual modalities, \texttt{No Touch} has the highest success rate drop for many of the variations (\textit{Peg/Hole Shape} and \textit{Sensor Noise} being the only exceptions) for both human-only and augmented demonstrations. On the other hand, \texttt{No Vision} has comparable (or sometimes even improved) \% success rate changes to \texttt{Full Model}, suggesting its reduced significance in our overall framework compared to the other modalities. Since our task begins immediately in a contact state that is maintained throughout a majority of the task's duration, it follows that F/T data provides the most valuable information about the task state. Proprioception may also give important task state information (as evidenced in the \% success rate change for the \texttt{No Prop.} model), as the position of the two end-effectors relative to each other is highly correlated to the position of the peg and hole relative to each other, which is essential knowledge in completing the insertion task. Thus, visual observations provide the least relevant information for our task while still being susceptible to many of the task variations. However, visual input may still be essential in task contexts outside of the one studied here, especially in situations with little to no F/T feedback (such as aligning the peg and hole objects to be in the same orientation before contact as was studied in our previous work \cite{ku2024evaluating}).

% \subsection{Attention Visualization}

% To gain further insight into the information being learned by our model, we  the attention weights in the latent vector cross-attention step of the PerceiverIO visuotactile encoder. We visualize the attention scores in cross-attention averaged over the 8 latent vectors for both visual (left and right wristview images) and tactile (force-torque history reading) in Figure \ref{fig:attention-visualizations}. We also plot the proportion of total attention for each modality during the course of a rollout.

% \begin{figure}[h]
%      \centering
%      \includegraphics[width=0.825\columnwidth]{images/‎ICRA_template_figure.jpeg}
%      \caption{\footnotesize{Attention visualization figure}}
%      \label{fig:attention-visualizations}
% \end{figure}

\subsection{Are these empirical trends also shown in the real world?}

\begin{figure}[h!]
     \centering
     \includegraphics[width=1.0\columnwidth]{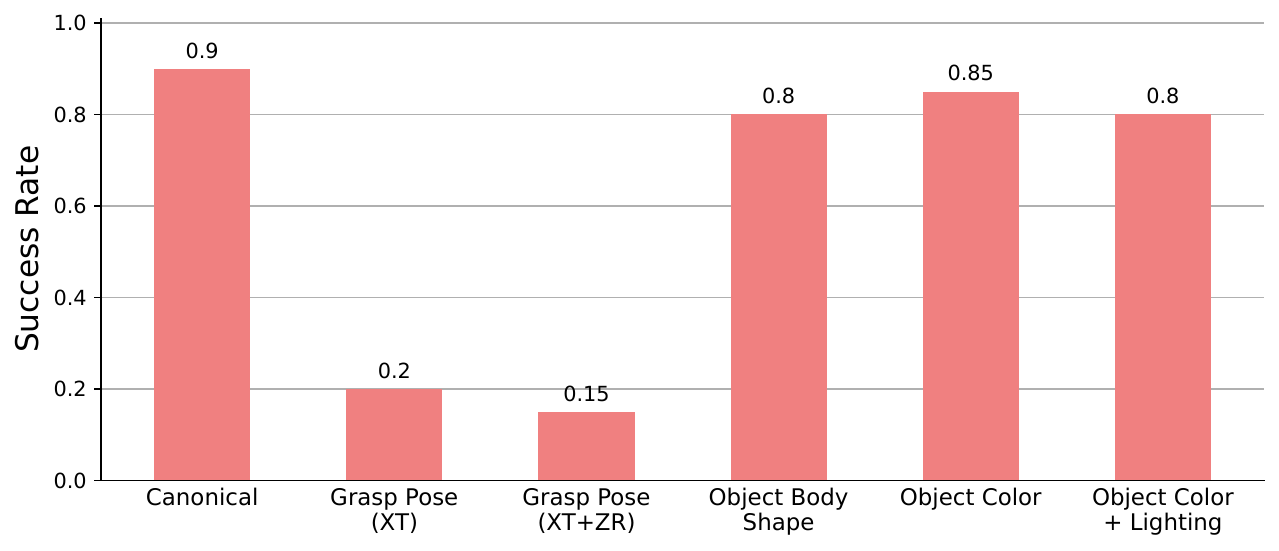}
     \caption{\footnotesize{Success rates on real-world task variations for a model trained on real-world non-augmented human demonstration data. The real-world model struggles the most with \textit{Grasp Pose} (both translation-only and translation with rotation) variations, aligning with our observations in simulation.}}
     \label{fig:real-world-canonical}
\end{figure}

As a real-world analog to our experiments in Section \ref{sec:05A_task-variation-difficulty}, we evaluate a real-world policy trained on a dataset of 50 human-generated demonstrations with no applied task variations on real-world versions of a subset of our task variations. Reported success rates over 20 rollouts can be found in Figure \ref{fig:real-world-canonical}.

\noindent \textbf{Takeaways:} Like in simulation, we observe that \textit{Grasp Pose} variations seem to be the most difficult to generalize to, while the model is able to handle the mostly unisensory perturbations of \textit{Object Body Shape} and \textit{Scene Appearance} (object color and lighting). We also notice that our model struggles with \textit{Grasp Pose} even when rotational grasp variations are removed; we hypothesize that this may be because a translational offset disrupts the desired behavior of lining up end-effector positions given from proprioceptive input in order to line up the peg and hole (i.e. solving the task can no longer be done by just matching the end-effector positions of the two arms). From these results, we believe that including \textit{Grasp Pose} variations into the training dataset (as was done in simulation through online augmentation) may also improve performance in the real world.

\begin{figure}[h!]
     \centering
     \includegraphics[width=1.0\columnwidth]{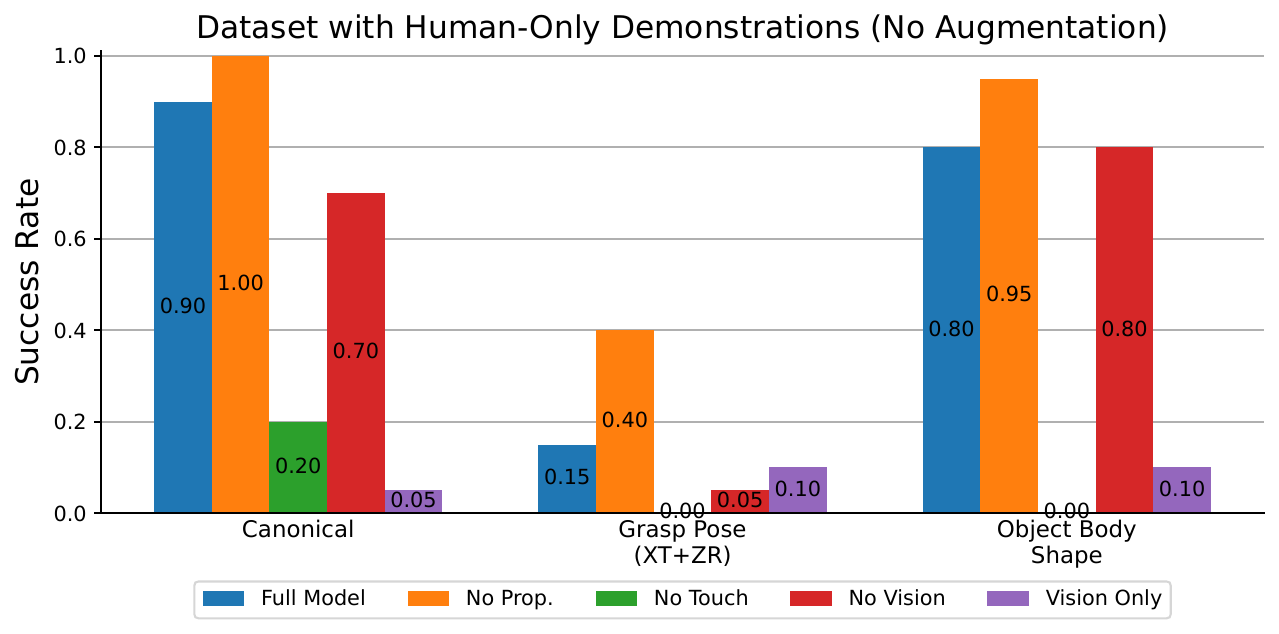}
     \caption{\footnotesize{Success rates on real-world task variations for models with different modality input combinations trained on no variations. The \texttt{No Touch} model sees the lowest success rates on all task variations, while the \texttt{No Prop.} model surprisingly sees an increase in performance over the \texttt{Full Model} for all task variations.}}
     \label{fig:real-world-modality-input-ablation}
\end{figure}

\textbf{Modality Input Ablation Study:} We conduct a reduced real-world analog to the ablation study in Section \ref{sec:05E_input-modality-combinations}. We train real-world policies on a dataset of only human demonstrations and evaluate them on a smaller subset of our real-world task variations. Reported success rates over 20 rollouts can be found in Figure \ref{fig:real-world-modality-input-ablation}.

\noindent \textbf{Takeaways:} Like in simulation, we observe that the removal of force-torque data as input (the \texttt{No Touch} model) leads to a significant drop in success rate for all task variations compared to the \texttt{Full Model}, including the no-variations \textit{Canonical} environment. We also see a small drop in performance for the \texttt{No Vision} model, somewhat aligning with our findings in simulation of the insignificance of visual input for our task. Surprisingly, we see performance increases in all task variations for the \texttt{No Prop.} model. We hypothesize that the small ranges of possible end-effector poses in our training dataset due to the high precision required for our task may cause our models to not learn much useful information from the proprioceptive embedding, though this observation may of course also be the result of a low sample size of trained models. Averaging the performance of models trained over multiple seeds (as was done in simulation) may give us some more robust results.

%% file: sections/07_conclusion.tex
\section{Conclusion}

% Real-world setup

% Behavior-cloning framework is extremely brittle and cannot recover from failure states, use a more sophisticated IL framework like diffusion policy and/or action chunking instead

Towards the goal of evaluating robustness of multisensory policies in contact-rich scenarios such as peg-in-hole object assembly, we present a pipeline for data collection, augmentation, policy training, and evaluation for an object assembly task across diverse observation-level task variations. Our experiments reveal that grasp variations pose the greatest challenge for generalization both in simulation and the real world, and incorporating them through data augmentation significantly improves performance on unseen variations. Additionally, we demonstrate that force-torque input is critical for robustness to task variations, while the removal of RGB input has minimal impact. For future work, we hope to expand our real-world experiments and validate online data augmentation as a method for increasing robustness in more realistic scenarios.
% We are able to execute known grasps in a controlled setting, and so we can set up a reliable trajectory replay pipeline that can perform online augmentation in the real world.} 

While we have demonstrated the ability of our system to learn the underlying task, we acknowledge that the behavior cloning setup used is highly susceptible to covariate shift and cannot recover from erroneous actions. We plan to extend our generalization studies using more advanced imitation learning frameworks, such as Diffusion Policy \cite{chi2024diffusionpolicy} and ACT \cite{zhao2023learning}, and compare their performance with our BC-MLP setup. Moreover, the constrained task initialization and action space in our setup highlight the need to explore more complex, longer-horizon tasks with broader action spaces, and to assess how observation-level task variations affect policies in these contexts.

% \kar{Limitation: Sim2Real is not done. Future work. }